\def\year{2022}\relax
\documentclass[letterpaper]{article} 
\usepackage{aaai22}  
\usepackage{times}  
\usepackage{helvet}  
\usepackage{courier}  
\usepackage[hyphens]{url}  
\usepackage{graphicx} 
\usepackage{natbib}  
\usepackage{caption} 
\DeclareCaptionStyle{ruled}{labelfont=normalfont,labelsep=colon,strut=off} 
\usepackage{algorithm}
\usepackage{algorithmic}

\usepackage{newfloat}
\usepackage{listings}
\floatstyle{ruled}
\newfloat{listing}{tb}{lst}{}
\floatname{listing}{Listing}
\usepackage{times}
\usepackage{latexsym}

\usepackage{microtype,todonotes}
\usepackage{enumitem}
\usepackage{amsthm}
\usepackage{bbm}
\usepackage{amsmath}
\usepackage{cleveref}
\usepackage{xspace}
\theoremstyle{definition}
\newtheorem{definition}{Definition}[section]

\newcommand{\anbn}{$a^nb^n$\xspace}
\newcommand{\dyck}{Dyck\xspace}
\newcommand{\wcwr}{$wcw^r$\xspace}
\newcommand{\wcwrn}{$(wcw^r)^n$\xspace}

\title{Learning Bounded Context-Free-Grammar via LSTM and the Transformer: \\Difference and Explanations}

\author{
    Hui Shi, \textsuperscript{\rm 1}
    Sicun Gao, \textsuperscript{\rm 1}
    Yuandong Tian, \textsuperscript{\rm 2}
    Xinyun Chen, \textsuperscript{\rm 3}
    Jishen Zhao\textsuperscript{\rm 1}
}
\affiliations{
    \textsuperscript{\rm 1}University of California San Diego, 
    \textsuperscript{\rm 2}Facebook AI Research, 
    \textsuperscript{\rm 3}University of California, Berkeley \\
    \{hshi, jzhao, sicung\}@ucsd.edu, yuandong@fb.com, xinyun.chen@berkeley.edu
}

\begin{document}
\maketitle

\begin{abstract}
Long Short-Term Memory (LSTM) and Transformers are two popular neural architectures used in natural language processing tasks. Theoretical results
show that both are 
Turing-complete and can represent any context-free languages (CFLs). In practice, it is often observed that 
the Transformer models have better representation power than the LSTM. But the reason is barely understood.
We study such practical differences between LSTM and the Transformer and propose an 
explanation based on their latent space decomposition patterns.
To achieve this goal, we introduce an oracle training paradigm, which forces the decomposition of the latent representation of LSTM and the Transformer, and supervises with the transitions of the corresponding Pushdown Automaton (PDA) of the CFL. With the forced decomposition, we show that the performance upper bounds of LSTM and the Transformer in learning CFL are close: both of them can simulate a stack and perform stack operation along with state transitions. However, the absence of forced decomposition leads to the failure of LSTM models to capture the stack and stack operations, while having a marginal impact on the Transformer model. 
Lastly, we connect the experiment on the prototypical PDA to a real-world parsing task to re-verify the conclusions. \footnote{The code is available at \url{https://github.com/shihui2010/learn_cfg_with_neural_network}}
\end{abstract}

\section{Introduction}
\label{sec:intro}
The LSTM network has achieved great success in various natural language processing (NLP) tasks \cite{sutskever2014sequence, wang2016attention}, and in recent years, the Transformer network keeps breaking the record of state-of-the-art performances established by LSTM-based models in translation \cite{vaswani2017attention}, question-answering \cite{devlin2018bert},  and so on\cite{dehghani2018universal, brown2020gpt3}. Besides exploring the capacity boundary of the Transformer network, there is an increasing interest in investigating the representation power of the Transformer network and explaining its advantage over the LSTM models theoretically and empirically. 

Existing analysis has proven that both LSTM \cite{siegelmann1995computational} and the Transformer network \cite{perez2019turing} are Turing-Complete. However, much empirical evidence shows both models are far from perfect in imitating even simple Turing machines \cite{dehghani2018universal, joulin2015inferring}. Several explanations of the performance gap between practice and theory are 1) some theoretical proofs relies on infinite precision assumption while the precision in practical computations is limited \cite{weiss2018practical, korsky2019computational}; 2) given fixed latent space dimension, according to the pigeonhole principle, as input sequence length goes towards infinity, there will be information that can not encode into the latent space or is forgotten \cite{hahn2020theoretical}. 

Despite the soundness of the theoretical proofs and explanations, none of them can directly explain the phenomenons in practice: 1) given the same computation precision, Transformer has an advantage over LSTM model in many cases; 2) both Transformer and LSTM fail even when the input sequence is short and their latent space dimension is huge. 

In this work, we study the empirical representation power of the LSTM and the Transformer networks and investigate the origination of their difference. We compare the models via learning context-free languages. The reason to study CFLs other than regular languages or Turing machines is that the learning of CFLs provides most insights into NLP tasks where understanding the underlying hierarchy of a sequence (\emph{e.g.} utterance and programs) is crucial. In the rest of the paper, we will first introduce an oracle training paradigm to predict the status of the PDA that accepts the CFL, and a regularizer to explicitly decompose the latent space of the LSTM and the Transformer such that the PDA state and the positions in the stack are encoded in distinct dimensions. Lastly, the experiment section exhibits the empirical results and leads us to the following conclusions:

\begin{itemize}[noitemsep,topsep=0pt]
    \item LSTM and the Transformer have a similar upper bound of empirical representation power in simulating PDAs. 
    \item LSTM fails to factorize its latent space to encode the state and multiple elements of the stack without explicit supervision, which is the pivot to its compromised performance in real-world tasks. Meanwhile, the Transformer is marginally affected by the absence of explicit decomposition regularization. 
    \item Language recognition is not a reliable task to compare the empirical capacity of LSTM and Transformer since the results are sensitive to the setting of PDAs, the hyperparameters of the models, and the training methods.
\end{itemize}


\section{Preliminaries and Definitions}
\label{sec:defs}
\begin{figure}[t]
    \centering
    \includegraphics[width=0.85\columnwidth]{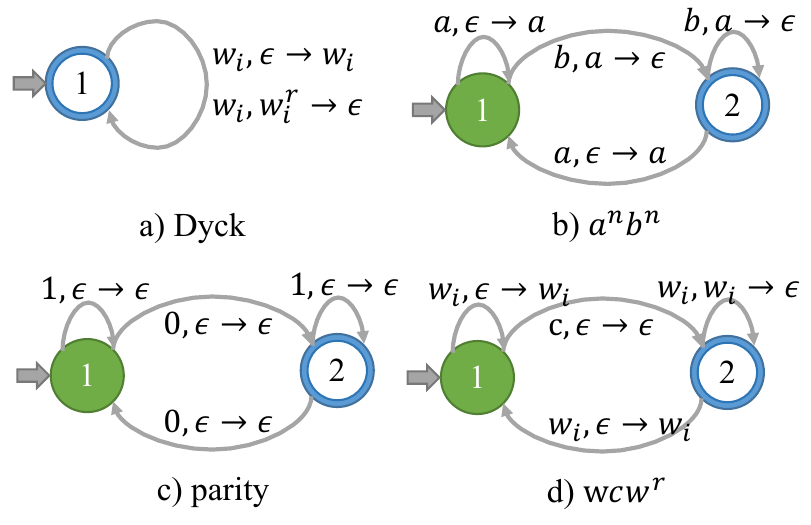}
    \caption{Graphical notations of PDAs for the CFLs.}
    \label{fig:automatons}
\end{figure}

\subsection{Context-Free language}
\label{ssec:cfg}
A context-free grammar $\mathcal{G}$ is defined by a collection of nonterminal symbols (\emph{i.e.} variables), terminal symbols (\emph{i.e.} the alphabet of $\mathcal{G}$), production rules and a start symbol $E$. The context-free language of $\mathcal{G}$ is the set of strings that can be derived from the start symbol by iteratively applying the production rules until there are no nonterminal symbols. 

A Pushdown Automaton (PDA) is a state machine with a stack that reads an input symbol and the top element in a stack at each step and performs the state transition and some stack operations (push/pop). Formally, a PDA can be defined as a tuple of $<Q, \Sigma, S, \delta, q_0, I, F>$. $Q$ is a finite set of states, and $q_0 \in Q$ is the initial state; $\Sigma$ is the alphabet of the inputs, $S$ is the set of stack symbols and $I \in S$ is the initial stack top symbol; $F \in Q$ is a set of accepting states. 
$\delta$ is a transition function $\delta(q \in Q, x \in (\Sigma \cup \{\epsilon\}), s \in S) \rightarrow q', s' $, where $\epsilon$ denotes an empty symbol and $s$ denotes the top element in the stack $S$. The transition function implies the stack operation. For example, let $*$ be a wildcard for arbitrary state or symbol, $\delta(*, \epsilon, *)$ represents transition that consumes no input symbols\footnote{For instance, the \textit{reduce} operation in shift-reduce parsing, as shown in Table~\ref{tab:shift_reduce}};
$\delta(*, *, \epsilon) \rightarrow *, s'$ where $s' \neq \epsilon$ is a stack push operation, and $\delta(*, *, s) \rightarrow *, \epsilon$ where $s \neq \epsilon$ is a stack pop operation. 

A PDA can be equivalently expressed by some CFG and vice versa \cite{schutzenberger1963context, chomsky1962context}. In learning CFGs, the neural networks are expected to learn the equivalent PDAs instead of the production rules. 

In this study, we are interested in bounded CFGs and PDAs where the recursion depth is finite and a stack with a finite size should be adequate for processing the CFLs. The reason is two-fold. First, we agree with the existing theoretical analysis that encoding an unbounded stack into finite space is the bottleneck for LSTM and Transformer, thus we focus on the investigation of their realistic representation power before they reach the theoretical upper bound (\emph{e.g.} number of recursions $\rightarrow \infty$). Second, in natural languages and even programming language, the nesting of the production rules are very limited (\emph{e.g.} $<100$), so it's important to understand LSTM and Transformer's behavior under bounded CFGs.

Four canonical PDAs (shown in Figure~\ref{fig:automatons}) are introduced as follows:

\begin{definition}[\dyck-(k, m)]
The language of paired parentheses of $k$ types up to $m$ recursions. \dyck is the prototypical language to any context-free languages \cite{kozen1997chomsky}. 
\end{definition}

\begin{definition}[\anbn]
 \anbn \ accepts and generates the set of strings that consists of $a$ and $b$, and every occurrence of n consecutive $a$s is followed by exact n successive $b$s. 
\end{definition}

\begin{definition}[Parity]
The binary strings that containing an even number of 0s. Parity can be expressed by a Deterministic Finite Machine (DFA). To formalize parity in PDA, we denote all parity transition function with no stack operation $\delta(*, *, \epsilon ) \rightarrow *, \epsilon$. 
\end{definition}

\begin{definition}[\wcwrn-(k, m)]
\wcwr-(k, m) generates strings that start with a sub-string $\omega$ followed by a character $c$, then followed by the reverse of $\omega$. The chars $w \in \omega$ come from a vocabulary $\Omega$ ($c \notin \Omega$) of size $k$. The string $\omega$ contains no more than $m$ chars. The \wcwrn-(k, m) generates strings that contains no more than $n$ substrings from \wcwr-(k, m). 
\end{definition}

\subsection{Long-short Term Memory network}
\label{ssec:lstm}
Given the input sequence embeddings $\mathbf{X}=\{x_t\}_{t=1}^n$ and initial hidden state and cell state $(h_0, c_0) \in \mathbf{R}^d$, LSTM \cite{lstm} produces the latent representation of the sequences $\{h_t\}_{t=1}^n$, where $h_t \in \mathbf{R}^d$, as follows:

\begin{equation}
\small
    \begin{aligned}
    & f_t = \sigma(\mathbf{W}_f h_{t-1} + \mathbf{U}_f x_t + b_f) \\
    & i_t = \sigma(\mathbf{W}_i h_{t-1} + \mathbf{U}_i x_t + b_i) \\ 
    & o_t = \sigma(\mathbf{W}_o h_{t-1} + \mathbf{U}_o x_t + b_o) \\
    & \Tilde{c}_t = tanh(\mathbf{W}_c h_{t-1} + \mathbf{U}_c x_t + b_c) \\
    & c_t = f_t \odot c_{t-1} + i_t \odot \Tilde{c}_t \\
    & h_t = o_t \odot tanh(c_t) 
    \end{aligned}
\end{equation}

\subsection{Transformer network}
\label{ssec:transformer}

The Transformer network \cite{vaswani2017attention} process the sequence via multi-head attention. Specifically, let the $att(Q, K, V)$ represent the scaled dot-product attention function over query $Q \in \mathbf{R}^{{r_0} \times r_{1}}$, key $K \in \mathbf{R}^{{r_0} \times r_{1}}$, and value $V \in \mathbf{R}^{{r_0} \times d}$, defined as:

\begin{equation}
\small
    att(Q, K, V) = softmax(\frac{Q K^\top}{\sqrt{r_1}} V)
\end{equation}

Given sequence $\mathbf{X}$, the Transformer encodes the sequences via multi-head attention followed by a position-wise feed-forward network (denote as $FFN(\cdot)$). 

\begin{equation}
\small
    \begin{aligned}
      & h_t = FFN([head_1;\cdots; head_k]); \\
      & head_i = att(y \mathbf{W}_i^Q, y \mathbf{W}_i^K, y_t \mathbf{W}_i^V)
    \end{aligned}
\end{equation}

To distinguish the order of input symbols, the $y_t$ is produced by summing a positional encoding with input encoding: 

\begin{equation}
    \small
    \begin{aligned}
    & p_{t, 2j} = sin(t/10000^{2j/r_1}) \\
    & p_{t, 2j + 1} = cos(t/10000^{2j/r_1}) \\
    & y_t = x_t + p_t
    \end{aligned}
\end{equation}

\begin{figure}
    \centering
    \includegraphics[width=\columnwidth]{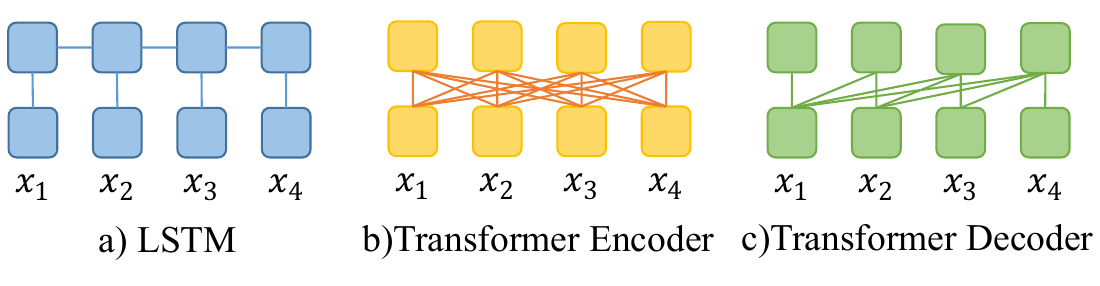}
    \caption{The information flow in LSTM, the Transformer encoder and decoder.}
    \label{fig:models}
\end{figure}

We distinguish the \textit{Transformer encoder} that allows attention between any pairs of input symbols and the \textit{Transformer decoder} that allows multi-head attention to compute only on past symbols. The comparison is illustrated in Figure~\ref{fig:models}.

\section{Oracle Training}
\label{sec:training}
In this section, we introduce the oracle training method assuming complete PDA transition steps are exposed to the model to provide the strongest supervision. 

Formally, given symbol sequences $\{X^{(i)}\}_{i=1}^n$ accepted by a PDA, the \textit{oracle} includes $\{X^{(i)}\}_{i=1}^n$, the states of DPA $\{S^{(i)}\}_{i=1}^n$ and the stack status $\{\mathcal{T}^{(i)}\}_{i=1}^n$ at each step while processing the symbol sequences. The oracle training forces the models to predict not only the next symbols as in the language model, but also the internal state and stack status. Hereinafter, for simplicity, we omit the superscript $i$ that denotes $i$-th sample. Let $X=\{x_t\}_{t=1}^\tau$, $S=\{s_t\}_{t=1}^\tau$, and $\mathcal{T}=\{T_t\}_{t=1}^\tau$, and the $T_t[j]$ being the $j$-th item in the stack at step $t$.

\begin{figure}
    \centering
    \includegraphics[width=\columnwidth]{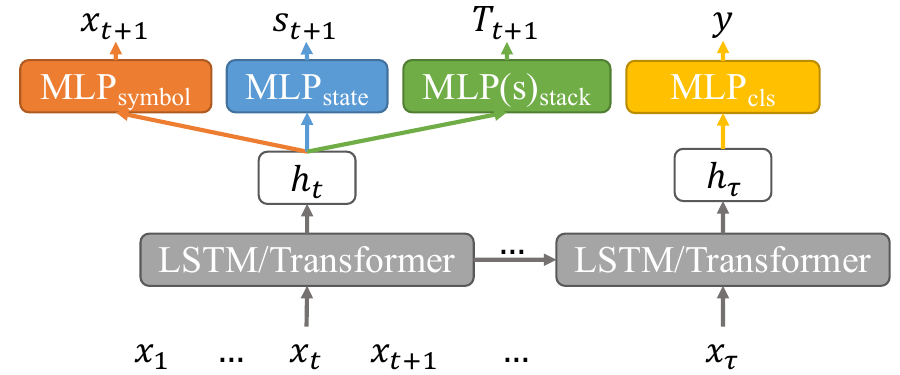}
    \caption{Oracle training.}
    \label{fig:oracle_training}
\end{figure}

Figure~\ref{fig:oracle_training} shows the generic architecture for oracle training. Several multi-layer perceptron (MLP) networks are employed to independently predict the symbol, state, and stack from the latent representation $h_t$. For stack status, a non-full stack is padded with the empty token $\epsilon$. Therefore, the models always predict a constant number of symbols for the stack and predict $\epsilon$s in the correct positions to indicate a non-full stack. We also include a language recognition task in which the model predicts if the sequences are accepted by the PDA. The language recognition task is widely used in arguing the inability of LSTM and Transformer in recognizing CFLs. Though from the PDA's perspective, the recognition is equivalent to the task of learning the transition, while in the experiment, we show both models benefit greatly from dense supervision in the oracle training, compared to the sparse supervision in the language recognition. 

\begin{figure}
\begin{minipage}{\columnwidth}
    \centering
    \includegraphics[width=\columnwidth]{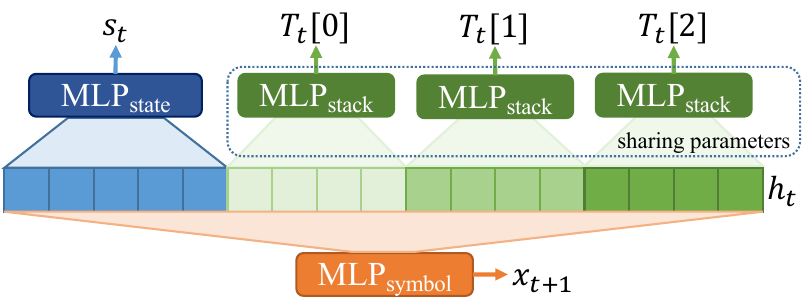}
    \caption{Forced decomposition of $h_t$.}
    \label{fig:forced_decomposition}
\end{minipage}

\begin{minipage}{\columnwidth}
    \centering
    \includegraphics[width=\columnwidth]{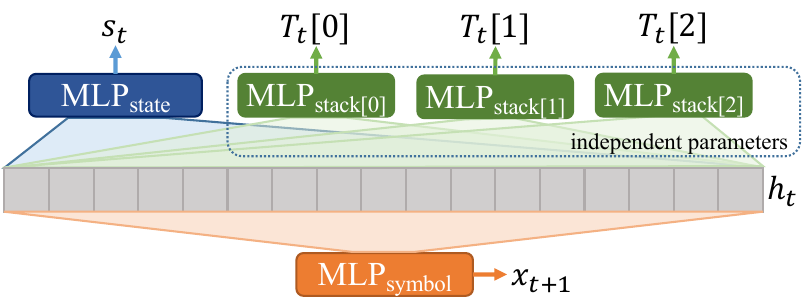}
    \caption{Latent decomposition of $h_t$ (with stack size of 3).}
    \label{fig:latent_decomposition}
\end{minipage}
\end{figure}

We also introduce two vital model configurations: \textbf{forced decomposition} (Figure~\ref{fig:forced_decomposition}) and \textbf{latent decomposition} (Figure~\ref{fig:latent_decomposition}). In forced decomposition, the latent representation vector $h_t$ is split into $m+1$ segments, where $m$ is the maximum stack size, to predict the PDA state and $m$ elements in the stack separately. Since each position in the stack shares the same set of stack symbols, we let the stack predictors $MLP_{stack}$ share parameters. In contrast, the latent decomposition uses whole $h_t$ but separate MLPs for prediction, and the $m$ stack predictors are independently trained. In both configurations, the next symbol is predicted based on the entire $h_t$ because the valid symbol for the next step depends on both the current state and stack.

Predictions of symbol, state, and stack are all trained with cross-entropy loss, and the three-part losses are summed
\footnote{An option is to assign weights to each term on the right hand side, our additional experiment in the Appendix shows that the choice of loss weights does not influence the main conclusion in the experiment. } : 

\begin{equation}
\small
\begin{aligned}
    & \mathcal{L}_{oracle} = \mathcal{L}_{symbol} + \mathcal{L}_{state} + \mathcal{L}_{stack}
\end{aligned}
\label{eq:loss}
\end{equation}

\section{Experiments}
\label{sec:exp}

\subsection{Canonical PDAs}
\label{ssec:pda_results}

\begin{figure*}[h]
\begin{minipage}{0.49\textwidth}
    \centering
    \includegraphics[width=\columnwidth]{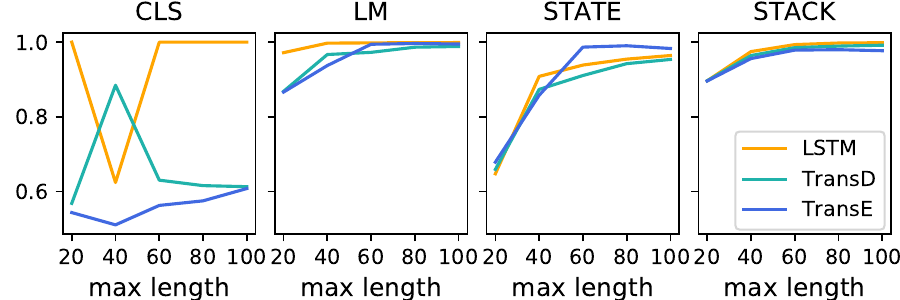}
    \caption{Performance on \anbn }
    \label{fig:anbn_s1_l1_forced}
\end{minipage}
\begin{minipage}{0.49\textwidth}
    \centering
    \includegraphics[width=\columnwidth]{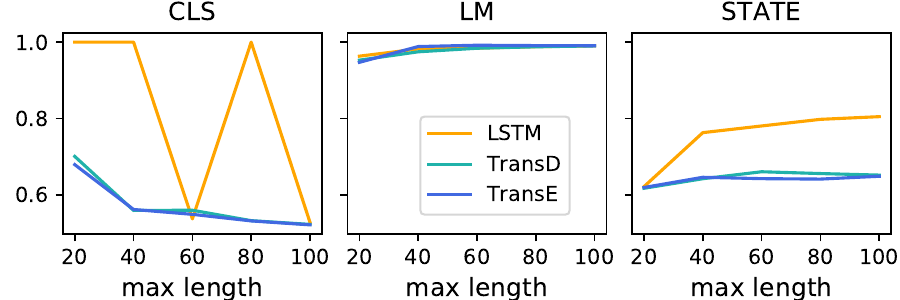}
    \caption{Performance on parity }
    \label{fig:parity_s1_l1_forced}
\end{minipage}
\end{figure*}

\begin{figure}[h]
    \centering
    \includegraphics[width=\columnwidth]{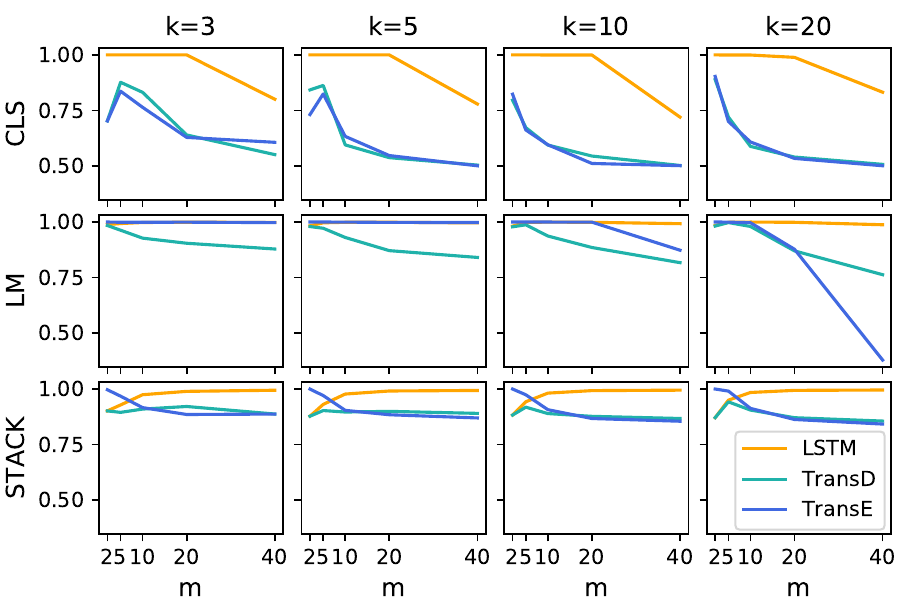}
    \caption{Performance on \dyck.  }
    \label{fig:dyck_s1_l1_forced}
\end{figure}

\begin{figure}[h]
    \centering
    \includegraphics[width=\columnwidth]{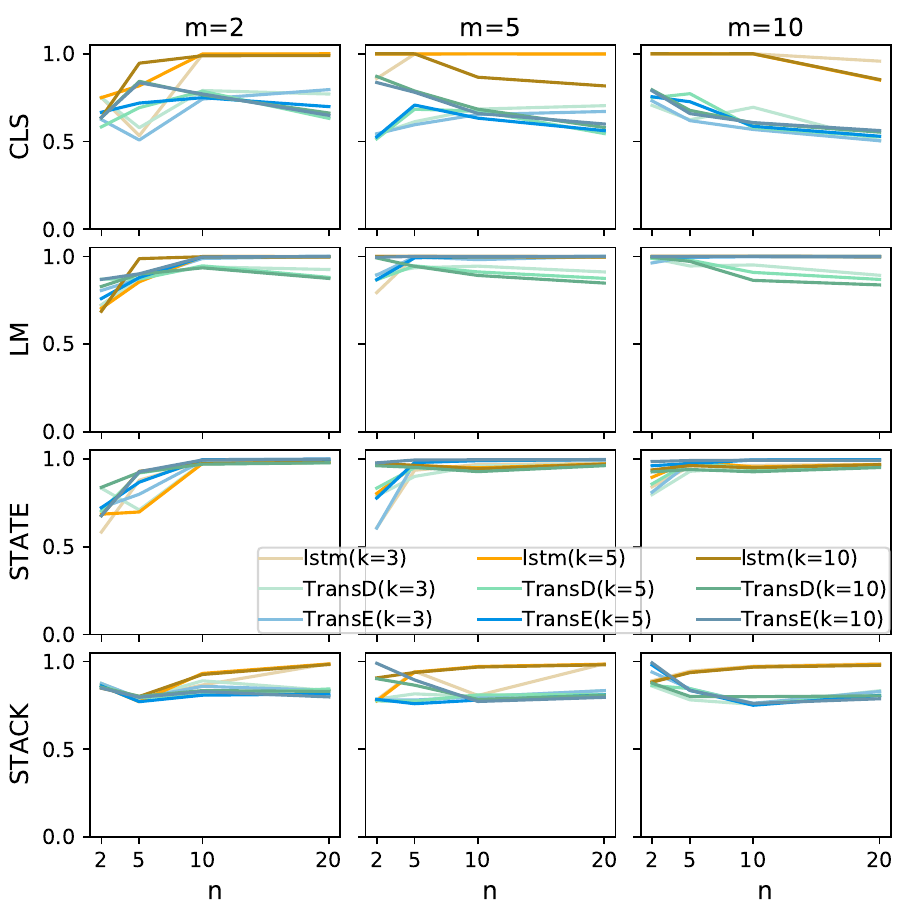}
    \caption{Performance on \wcwrn.  }
    \label{fig:wcwr_perf}
\end{figure}

In this section, we evaluate the representation power of LSTM and Transformer by simulating canonical PDAs. 

\textbf{Data generation. } For the PDAs introduced, we generate the 50k sequences for training and another 50k for testing for each configurations (\emph{i.e.} $\{k, m, n\}$ when applicable). Exceptionally, for \wcwr(n=2, m=2, k=*) we enumerate all accepted sequences since the set of accepted sequences are limited. For each sequence, the ground-truth PDA annotates the sequence and produces the state and stack labels for oracle training. Meanwhile, a corrupted sequence that is not accepted by the PDA is generated for the language recognition (classification) task. For language modeling, we compute the valid symbols at each step $t$ given sequence $x_1,\cdots, x_{t-1}$ and denote this validity over alphabet as \textit{LM mask}. 

\textbf{Model configurations.} For each PDA task, we set the hidden size in the LSTM model and the latent dimension in the Transformer model to be $\alpha * (|Q| + m * |S|)$, where $\alpha$ is the scale factor, $|Q|$ is the size of states, $m$ is the maximum recursion (\emph{i.e.} maximum stack size), and $|S|$ is the size of stack symbols\footnote{For PDAs introduced, we set $S=\Sigma \bigcup \{I, \epsilon\}$}. We always let $\alpha \ge 1$ such that the latent vector $h_t$ will always have enough dimensions to encode both the state and stack. For the Transformer encoder and decoder, the number of attention heads is 8. For all models, unless specified otherwise, the number of layers is 1 and $\alpha=1$. We set all MLP modules to be a two-layer feed-forward network with a sigmoid as an activation function. The hidden dimension of the MLPs is twice their input feature dimension. For both models, we use an embedding layer to encode input symbols to $\mathbf{R}^{2|\Sigma|}$. For the Transformer model, the filter size for the position-wise feed-forward network is 32, and we apply a dropout layer with a rate of $0.1$. 

\textbf{Training. } Models are trained with Adam optimizer with a learning rate of 0.001 on the AWS platform. The models are trained for 200 epochs or up to convergence. We introduce \textit{two-phase training} for language recognition tasks, \emph{i.e.} classify whether the sequences are accepted by the PDA. In phase 0, the models are initialized and solely trained on a classification task, while in phase 1, the models are re-trained on classification tasks after being trained with oracle training.

\textbf{Metrics. } 1) Classification accuracy: portion of the sequences that are correctly accepted/rejected. 2) LM accuracy: percentage of predicted symbols that are valid according to the LM mask. 3) State accuracy: accuracy in predicting the current PDA state. 4) Stack accuracy: accuracy in predicting current stack status, which is the average prediction accuracy over each stack position (including empty positions).

\textbf{Overall results. } Figure~\ref{fig:anbn_s1_l1_forced},~\ref{fig:parity_s1_l1_forced},~\ref{fig:dyck_s1_l1_forced},~\ref{fig:wcwr_perf} shows the LSTM and Transformer performance over multiple configuration of PDAs \footnote{These figs shows phase 0 classification accuracy}. There are few observations from the results: 1) LSTM has higher accuracy in learning PDAs, as it generally achieves higher accuracy in both state and stack predictions, especially shown in Figure~\ref{fig:parity_s1_l1_forced},~\ref{fig:wcwr_perf}. The results shows that when CFGs are bounded, LSTM does not need external memory to simulate the stack. 2) For predicting the state, Transformer decoder behaves in a similar way to LSTM, as shown in state accuracy in Figure~\ref{fig:anbn_s1_l1_forced},~\ref{fig:wcwr_perf}. In \anbn and \wcwrn, the Transformer encoder slightly outperforms LSTM and Transformer decoder. This indicates that though past information should be adequate to determine the current status of PDA, it's still beneficial to the neural models to foresee the future sequences. 3) Both language recognition task and language modeling solely should not be used to examine the capability of neural models in learning CFG, since they do not fully reflect the capability of models to learn the internal dynamics in state transition and stack operation (Figure~\ref{fig:parity_s1_l1_forced}), and reversely learning the precise PDA does not necessarily lead to perfection in language recognition and language modeling (Figure~\ref{fig:dyck_s1_l1_forced}, \ref{fig:wcwr_perf}). 

\begin{figure}[t]
    \centering
    \includegraphics[width=\columnwidth]{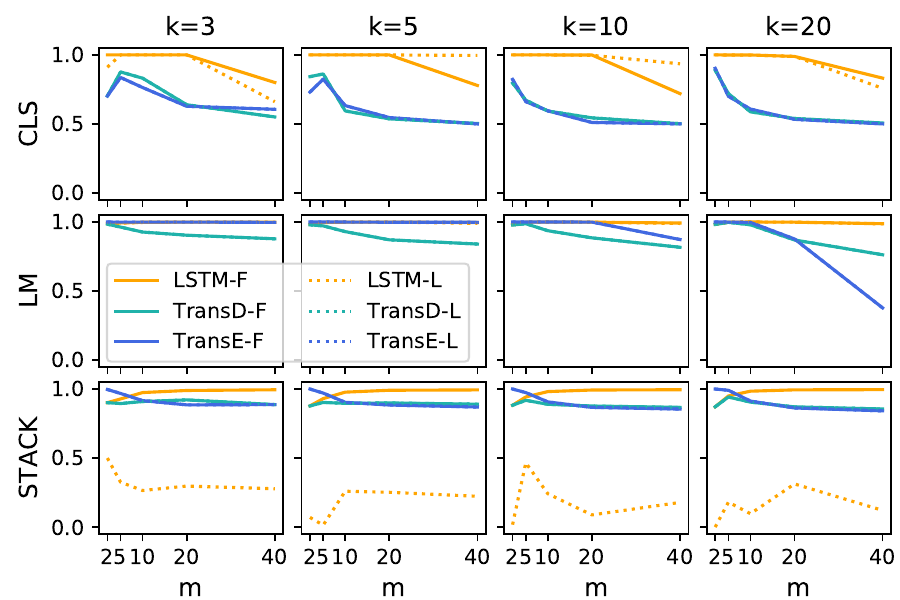}
    \caption{Performance on \dyck with forced (solid lines) and latent (dotted lines) decomposition. }
    \label{fig:dyck_decom}
\end{figure}

\begin{figure}[t]
    \centering
    \includegraphics[width=\columnwidth]{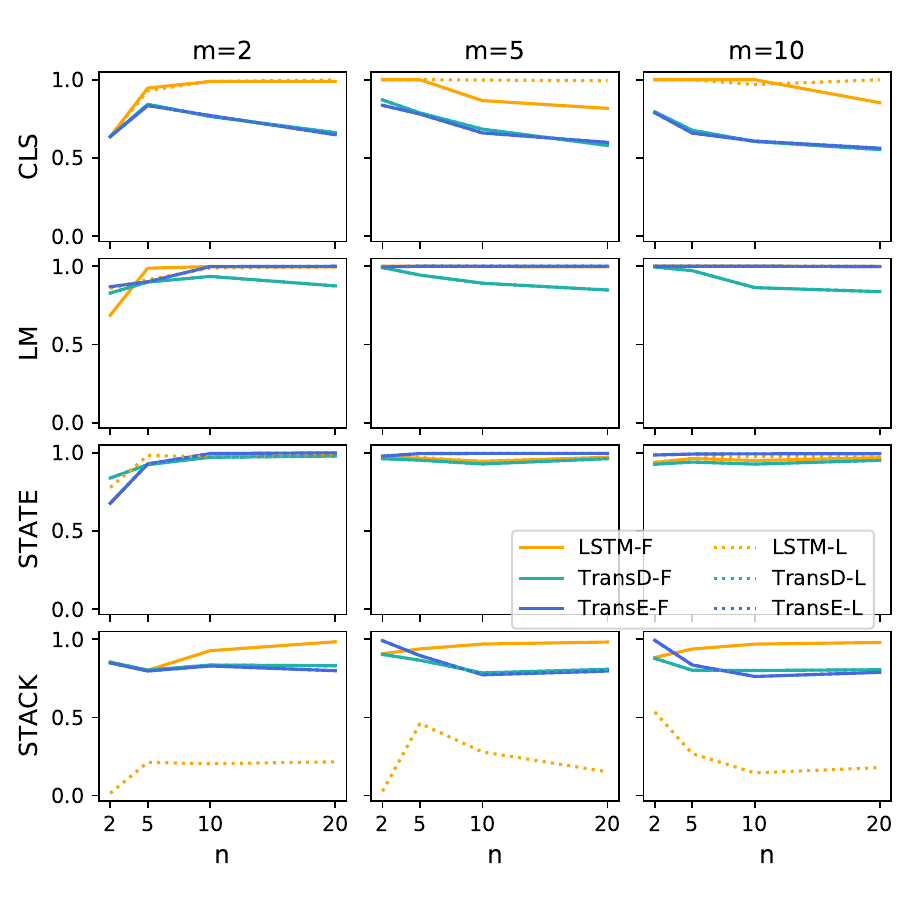}
    \caption{Performance on \wcwrn with forced (solid lines) and latent (dotted lines) decomposition ($k=10$). }
    \label{fig:wcwr_decom}
\end{figure}

\begin{figure}
    \centering
    \begin{minipage}{0.49\columnwidth}
    \centering
    \includegraphics[width=\columnwidth]{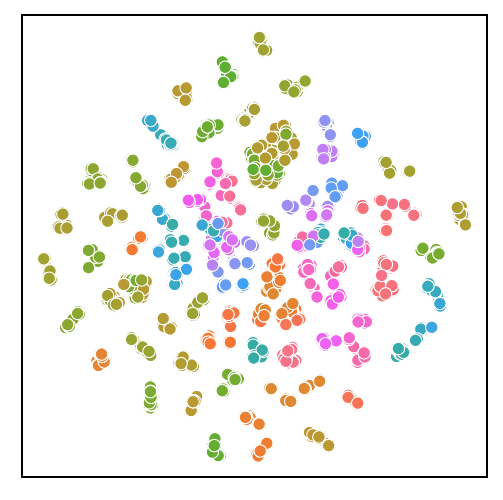}
    a) Forced 
    \end{minipage}
    \begin{minipage}{0.49\columnwidth}
    \centering
    \includegraphics[width=\columnwidth]{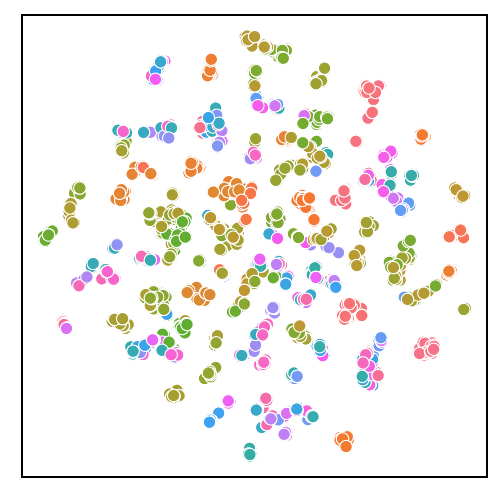}
    b) Latent 
    \end{minipage}
    \caption{t-SNE analysis on LSTM hidden states trained on \dyck(k=3,m=5). }
    \label{fig:tsne}
\end{figure}

\textbf{Decomposition hardship of LSTM.} Though LSTM and Transformer show comparable representation power under forced decomposition setting, the disadvantage of LSTM in the latent decomposition training is significant, as can be viewed from Figure~\ref{fig:dyck_decom}, \ref{fig:wcwr_decom}, meanwhile, Transformer models are roughly as good as they are trained with forced decomposition. The failure of LSTM in stack prediction is key to its disadvantage in many tasks \cite{devlin2018bert}. Figure~\ref{fig:tsne} shows the two-component t-SNE \cite{tsne} results of the LSTM hidden states $h_t$. The different colors represent distinct stack status in the oracle. As shown, the hidden states in forced decomposition tend to separately encode different stack status, while in the latent decomposition, there are much more clusters containing multiple colors. The mixed-color clusters indicate the failure of stack predictor to correctly predict the stack status, and also prevents LSTM itself to learn the correct stack operations.  

\begin{figure}
    \centering
    \includegraphics[width=\columnwidth]{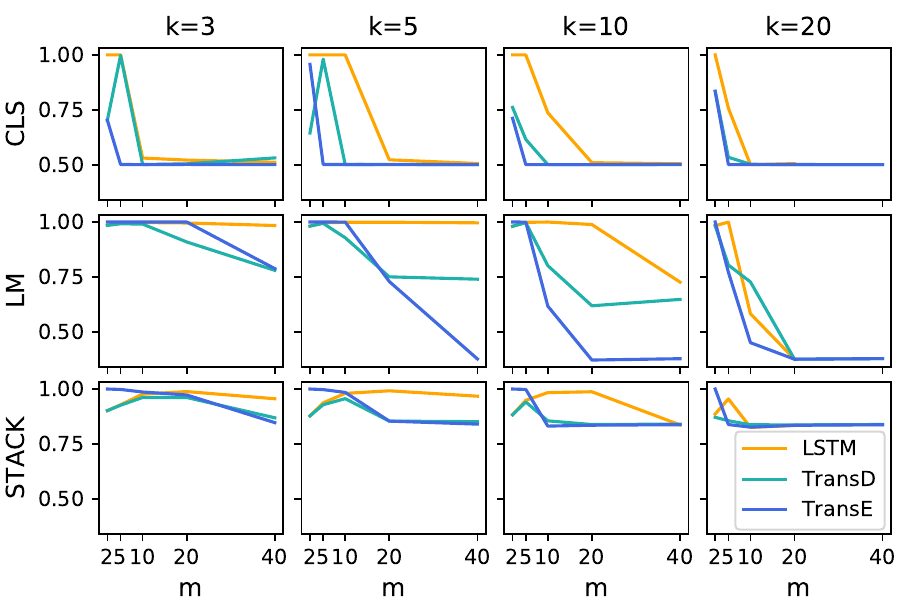}
    \caption{Performance of four-layer models on \dyck with $\alpha=4$}
    \label{fig:dyck_scale4}
\end{figure}

\textbf{Scale factor $\alpha$. } Generally, a larger number of parameters brings higher representation power. We verified the conclusion that LSTM and Transformer have similar representation power of learning CFG on larger models. Figure~\ref{fig:dyck_scale4} shows the performance of four-layer LSTM and Transformer model with scaling factor $\alpha = 4$. Viewing from the stack prediction accuracy, the conclusion still holds. We also examined the conclusion on factorization on larger scale models shown in Figure~\ref{fig:dyck_decom_4}. The full results of the larger models can be found in the Appendix. 

\begin{figure}[!t]
    \centering
    \includegraphics[width=\columnwidth]{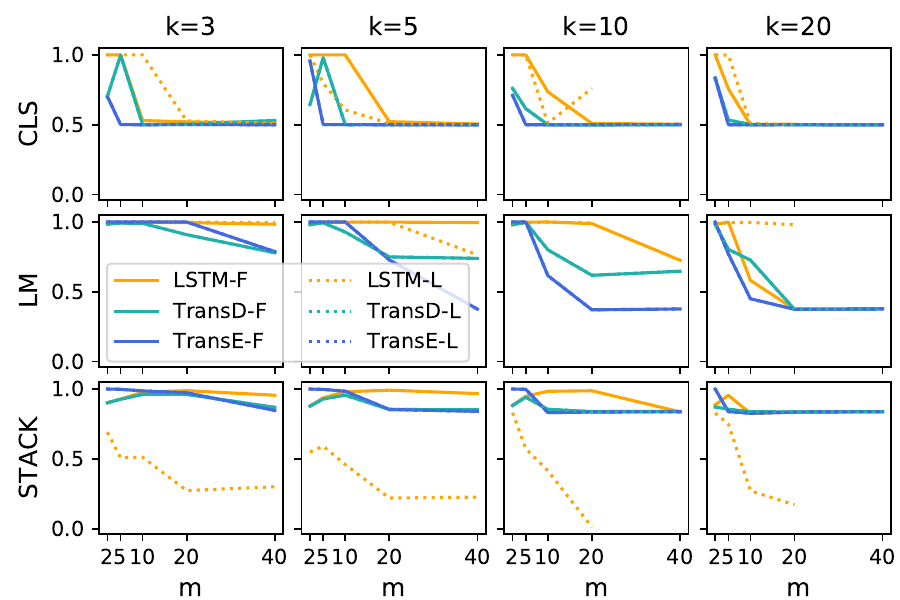}
    \caption{Four-layer models ($\alpha=4$) on \dyck with forced (solid lines) and latent (dotted lines) decomposition.}
    \label{fig:dyck_decom_4}
\end{figure}

\begin{figure}[t]
    \centering
    \includegraphics[width=\columnwidth]{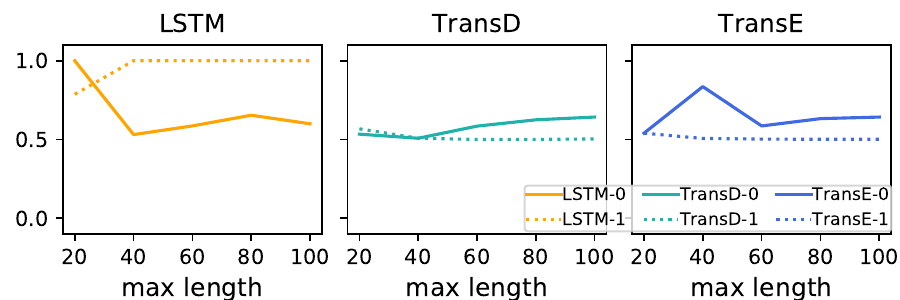}
    \caption{Two-phase classification accuracy on \anbn. }
    \label{fig:anbn_cls}
\end{figure}

\begin{figure}[t]
    \centering
    \includegraphics[width=\columnwidth]{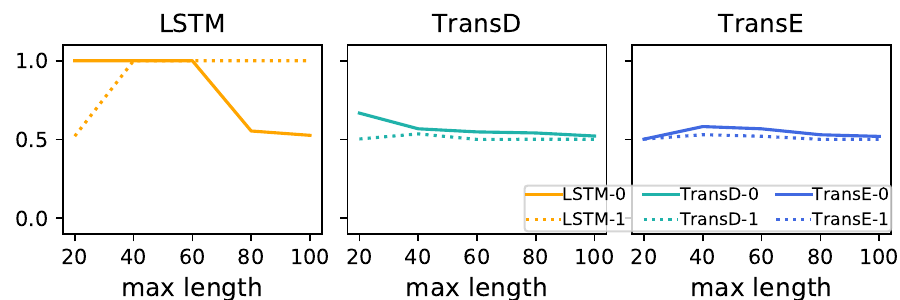}
    \caption{Two-phase classification accuracy on parity. }
    \label{fig:parity_cls}
\end{figure}

\begin{figure}[t]
    \centering
    \includegraphics[width=\columnwidth]{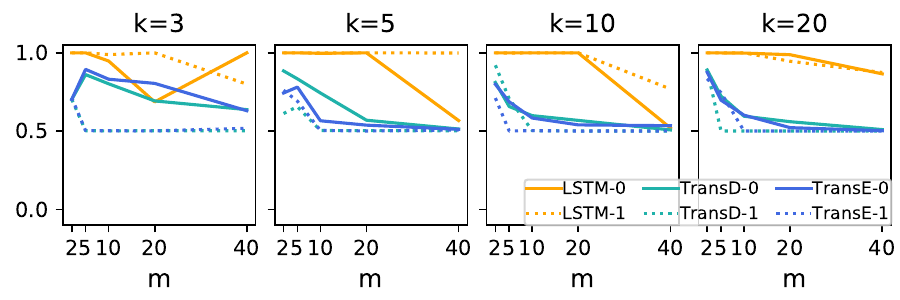}
    \caption{Two-phase classification accuracy on \dyck. }
    \label{fig:dyck_cls}
\end{figure}

\textbf{Why language recognition accuracy is not a good indicator. } We detail the four-tier reasons for our earlier conclusion that language recognition accuracy is unfair in judging the models' capability of learning PDA. Firstly, we explain the observations from Figure~\ref{fig:dyck_s1_l1_forced}, \ref{fig:wcwr_perf} that perfection in state and stack prediction does not lead to accurate recognition: the language recognition requires a higher level of reasoning than just learning a set of PDA transition functions. A rejected sequence for PDA might be due to the current symbol not being accepted given the state and stack top, popping an empty stack, or exceeding the maximum recursion. Thus the classification decision boundary might be inseparable for an MLP without special design. Furthermore, for any models, the error message might occur at random steps, and the models have to preserve and pass the message to the last step for the classification prediction in the general training paradigm for language recognition. This is mainly why the classification accuracy of LSTM declines much more abruptly than of the Transformers since Transformer can pass information between any pairs of inputs, thus the comparison of LSTM and Transformer will be sensitive to sequence lengths. Besides sequence length, the model size also influences the recognition accuracy differently. Comparing the classification accuracy of Figure~\ref{fig:dyck_s1_l1_forced} and Figure~\ref{fig:dyck_scale4}, the relative advantage of two models are reversed when models scale up. Lastly, Transformer models may recognize language in some manners that are dissimilar to PDAs. To prove this, we show the classification accuracy of phase 0 and phase 1 in  Figure~\ref{fig:anbn_cls},~\ref{fig:parity_cls},~\ref{fig:dyck_cls}. LSTM generally improves after oracle training (phase 1), especially in \dyck (Figure~\ref{fig:dyck_cls}), while the Transformer models suffers from oracle training (Figure~\ref{fig:anbn_cls}-\ref{fig:parity_cls}).

\begin{table*}[ht]
\begin{minipage}{0.38\textwidth}
    \centering
    \resizebox{\columnwidth}{!}{
    \begin{tabular}{rcl}
       C  & := & CP S $|$ CP V $|$ S \\
       CP  & := &  S and $|$ S after\\
       S  & := & V $|$ V twice $|$ V thrice \\
       V  & := & VP D $|$ VP left $|$ VP right $|$ D \\
       VP  & := & D opposite $|$ D around \\
       D  & := & U $|$ U left $|$ U right \\
       U  & := & walk $|$ look $|$ run $|$ jump $|$ turn \\
    \end{tabular}
    }
    \caption{Domain Specific Language for SCAN linguistic commands. }
    \label{tab:scan_dsl}
\end{minipage}
\begin{minipage}{0.6\textwidth}
    \centering
    \resizebox{\columnwidth}{!}{
    \begin{tabular}{c|c|c|c}
    \hline\hline
      step & stack & unconsumed symbols & transition $\delta$ \\
      \hline
      0 & [] & jump left and turn opposite left & -- \\
      1 & [U] & left and turn opposite left & $\delta(left, \epsilon) \rightarrow U$ \\
      2 & [D] & and turn opposite left & $\delta(and, U) \rightarrow D$ \\
      3 & [V] & and turn opposite left & $\delta(\epsilon, D) \rightarrow V$ \\
      4 & [S] & and turn opposite left & $\delta(\epsilon, V) \rightarrow S$ \\
      5 & [CP] & turn opposite left & $\delta(and, S) \rightarrow CP$ \\
      6 & [CP, D] & opposite left & $\delta(turn, \epsilon) \rightarrow U$ \\
      7 & [CP, VP] & left & $\delta(opposite, D) \rightarrow VP$ \\
      8 & [CP, V] &  & $\delta(left, VP) \rightarrow V$ \\
      9 & [C] &  & $\delta(\epsilon, [CP, V]) \rightarrow C$ \\
      \hline
      \multicolumn{4}{c}{Padded Sequence} \\
      \multicolumn{4}{c}{[jump, left, and, $<$reduce$>$, $<$reduce$>$, turn, opposite, left, $<$reduce$>$]}\\
      \hline
    \end{tabular}
    }
    \caption{Shift-reduce parsing of SCAN commands. }
    \label{tab:shift_reduce}
\end{minipage}
\end{table*}

\subsection{Shift-Reduce Parsing}
\label{ssec:scan_results}

\textbf{Dataset. } SCAN \cite{scan} is a semantic parsing dataset consisting of commands in natural language and sequences of actions to execute the commands. Tab.~\ref{tab:scan_dsl} shows the generation rules producing SCAN commands. The dataset contains 16728 training samples and 16728 test samples. In our experiment, instead of parsing to the sequence of actions, we parse the linguistic command according to its CFG production rules (Tab.~\ref{tab:scan_dsl}) using shift-reduce parsing. Since there are production rules in the CFG that map a single nonterminal variable to another one, the corresponding PDA is nondeterministic. To facilitate the training, we insert a special token \textit{<reduce>} to make the process deterministic. Tab.~\ref{tab:shift_reduce} illustrates an example of the annotation of the sequence with stack status and the padding process to insert \textit{<reduce>} tokens for reduction operations that do not consume symbol from the input sequence. The equivalent PDA of SCAN's CFG has only one state $q_0$. Alphabet $\Sigma$ covers English words in the linguistic commands and the \textit{<reduce>} token, and the stack symbols are $\{C, CP, S, V, VP, D, U\}$. 

\textbf{Models and metrics. } The model configurations are the same as in learning the canonical PDAs. We compute perplexity to evaluate language modeling and compute the accuracy of the stack prediction as the parsing accuracy. 

\textbf{Results. } Table~\ref{tab:scan_results} sums up the results of LSTM and Transformer on SCAN dataset, which is consistent with the observations in previous section: LSTM and Transformer decoder perform similarly in language modeling and parsing when decomposition is forced, and the parsing accuracy of LSTM suffers heavily from latent decomposition setting. The Transformer encoder can see the whole sequence at each step, so it achieves an almost lower bound of perplexity and has the highest parsing accuracy. 

\begin{table}[h]
    \centering
    \resizebox{\columnwidth}{!}{
    \begin{tabular}{c|cc}
    \hline
       Model  & Perplexity & Accuracy \\
    \hline
        LSTM($\alpha=1$, Forced) & 2.705 & 91.30 \\
        LSTM($\alpha=1$, Latent) & 2.705 & 16.61 \\
        LSTM($\alpha=4$, Forced) & 2.706 & 91.30 \\
        LSTM($\alpha=4$, Latent) & 2.708 & 35.00 \\
    \hline 
        Transformer(D, $\alpha=1$, Forced) & 2.710 & 91.02 \\
        Transformer(D, $\alpha=1$, Latent) & 2.713 & 76.61 \\
        Transformer(D, $\alpha=4$, Forced) & 2.708 & 91.30 \\
        Transformer(D, $\alpha=4$, Latent) & 2.710 & 90.56 \\
    \hline
        Transformer(E, $\alpha=1$, Forced) & 1.020 & 99.84 \\
        Transformer(E, $\alpha=1$, Latent) & 1.014 & 72.39 \\
        Transformer(E, $\alpha=4$, Forced) & 1.002 & 99.99 \\
        Transformer(E, $\alpha=4$, Latent) & 1.001 & 99.29 \\
    \hline
    \end{tabular}
    }
    \caption{Perplexity and parsing accuracy on SCAN. \textbf{E} denotes Transformer encoder and \textbf{D} denotes decoder.}
    \label{tab:scan_results}
\end{table}

\section{Related Works}
\label{sec:related}
There are many theoretical analyses on the representation power of LSTM and the Transformer and their comparisons that motivate this work to re-examine their representation power from an empirical aspect. \citet{siegelmann1995computational} firstly established the theory that given infinite precision and adequate number of hidden units RNNs are Turing-Complete, and \citet{holldobler1997designing} has designed an one-unit vanilla RNN counter for recognizing counter languages (\emph{e.g.} $a^nb^n$ and $a^nb^nc^n$) with finite range of $n$. Recently, \citet{hewitt2020rnns} proposed the construction of RNN that performs stack operations as in PDA and encodes PDA stack within hidden states of RNN without external memory. \citet{perez2019turing} proofs that with arbitrary precision, the Transformer network could simulate the single execution step for Turing machine, then by induction the Transformer is Turing-Complete. The majority of theoretical analyses emphasize that limited computation precision may break the proofs and compromise the performance in practice. Nevertheless, the models are not supervised with either step-by-step execution of the Turing Machine or the actual counters, which might be the crux to the failures of both models in reality. 

This work also closely relates to previous attempts to connect LSTM and transformer model with a specific type of languages, \emph{e.g.} languages from Chomsky's hierarchy \cite{chomsky1956three} and counter languages. 1) For regular languages (representable by DFAs),  \citet{michalenko2019representing} shows the empirical ability of LSTM to represent DFAs and \citet{rabusseau2019connecting} has proposed a construction method of RNNs from a weighted DFA. 2) For context-free languages (representable by PDAs), \citet{sennhauser2018evaluating} observed that CFGs are hardly learnable by LSTM models. on the other hand, \citet{bhattamishra2020practical} showed LSTM could learn CFGs with bounded recursion depth but the performance will be limited for infinite recursion. 3) For counter languages, the Transformer network \citep{bhattamishra2020ability} and LSTM \citet{suzgun2019lstm} have been trained to predict the outputs of a dynamic counter. However, their results disagree on the capability of LSTM in representing DYCK languages. In our work, we focus on bounded CFG since the capacity of learning regular languages is widely agreed upon while there are disputes on the CFG level. 

Finally, observing the defects of both LSTM and Transformer in learning CFG and algorithmic tasks, many works propose to use external memory to enhance the LSTM model \cite{joulin2015inferring, das1992learning, suzgun2019memory}, introduce recurrence in the Transformer network \cite{dehghani2018universal}, and design specialized architectures \cite{graves2014neural, hao2018context, sukhbaatar2015end, stogin2020provably}. Though it's commonly believed that LSTM with finite memory, \emph{i.e.} hidden states, can not handle CFGs which requires infinite stack spaces, we investigate the capacity of LSTM and transformer in bounded CFGs that requires finite-size stack and conclude that finite memory is not the bottleneck of LSTM capacity in learning CFGs.  

\section{Conclusion}
\label{sec:conclusion}
We illustrate that only the state and stack prediction accuracy trained with dense supervision and explicit decomposition regularizer are the fair and stable metric to compare the empirical representation power of LSTM and the Transformer network. Then we conclude that both LSTM and Transformer network can simulate context-free languages with bounded recursion with a similar representation power, and unveiled the disadvantage of the LSTM model in practice is from its inability to decompose the latent representation space. 

\bibliography{ref.bib}

\appendix
\begin{figure}[t]
\centering
\begin{minipage}[t]{0.87\columnwidth}
    \centering
    \includegraphics[width=\columnwidth]{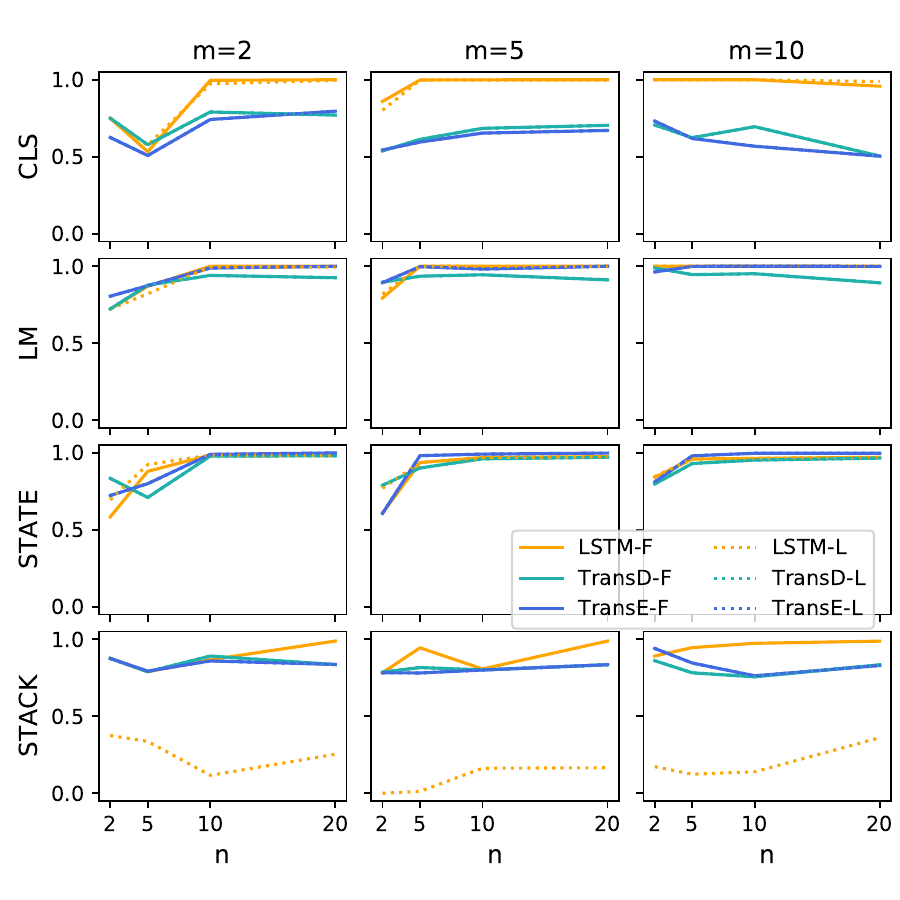}
\end{minipage}
\begin{minipage}[b]{0.87\columnwidth}
    \centering
    \includegraphics[width=\columnwidth]{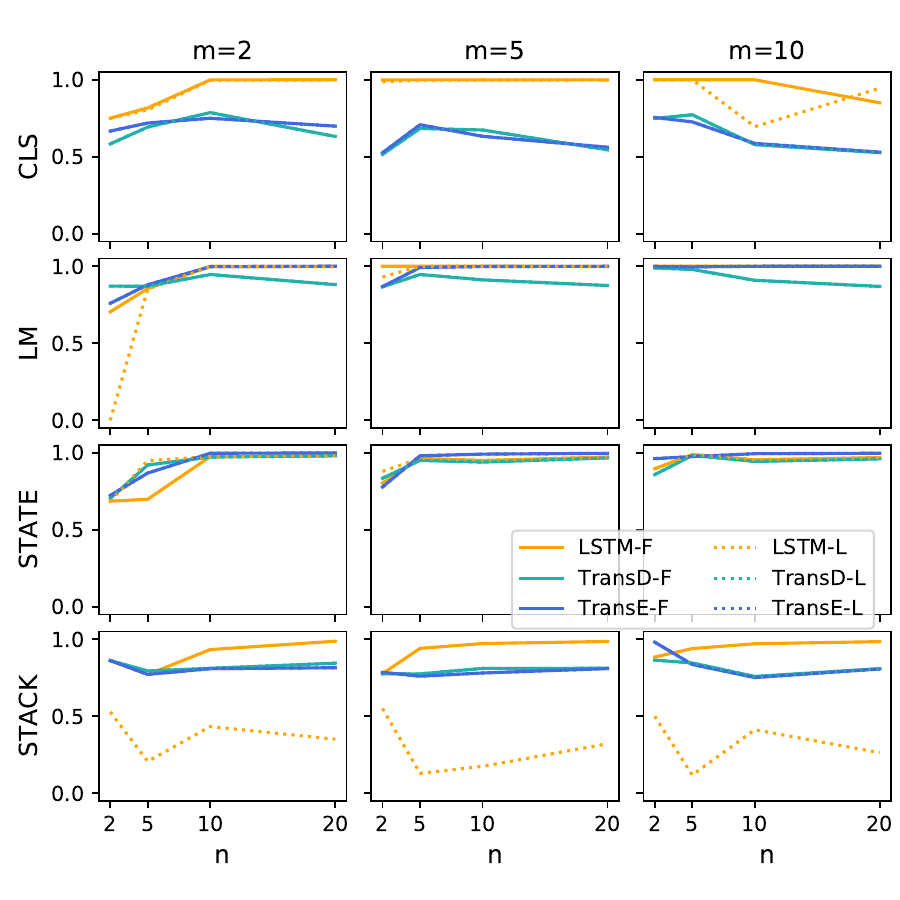}
\end{minipage}
    \caption{Performance on (\wcwr)$^n$. Upper figure shows results with $k=3$, and the lower shows results with $k=5$.}
    \label{fig:wcwr_decom_s1_app}
\end{figure}

\section{Additional Factorization Results}
\begin{figure}
    \centering
    \includegraphics[width=\columnwidth]{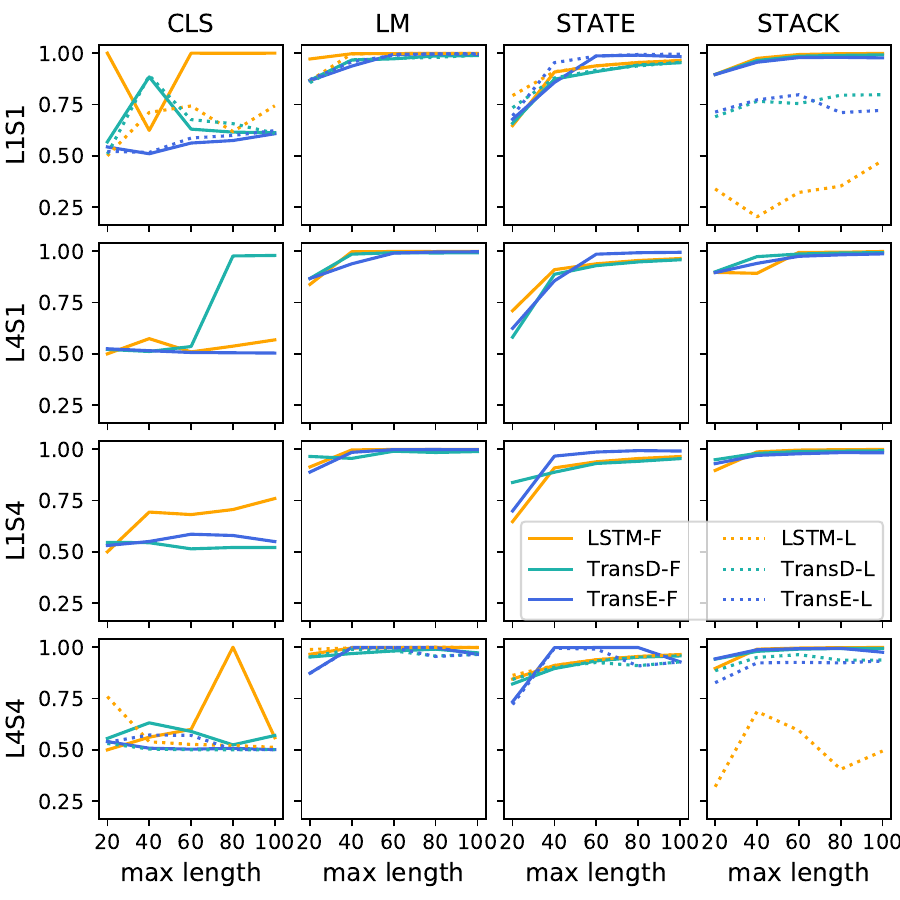}
    \caption{Performance on \anbn. Y-label shows number of layers and scaling factor $\alpha$. For example, \textit{L1S4} represents single-layer model with $\alpha=4$. }
    \label{fig:anbn_model_scale}
\end{figure}

Figure~\ref{fig:wcwr_decom_s1_app},~\ref{fig:anbn_model_scale},~\ref{fig:wcwr_decom_s4} show additional results for factorization. Unless specified, the model are configured with a single layer and $\alpha=1$. All the results support the conclusion that LSTM without forced factorization hardly learns to simulate the PDA stack properly. 

\section{Sensitivity to Loss Weights}
\textbf{Loss function:} In Equation~(\ref{eq:loss}), the weights of $\mathcal{L}_{symbol}$, 
$\mathcal{L}_{state}$, $\mathcal{L}_{stack}$ are set to 1. To confirm the conclusion is general without dependent on the choice of weight values, we examine the model performance using the learnable loss weights \cite{kendall2018multi}, and the oracle training loss has the form of:

\begin{equation}
\small
\begin{aligned}
    & \mathcal{L}_{oracle} = \frac{1}{2 \sigma_1^2} \mathcal{L}_{symbol} + \frac{1}{2 \sigma_2^2}\mathcal{L}_{state} \\ 
    & \quad  + \frac{1}{2 \sigma_3^2} \mathcal{L}_{stack} + log(\sigma_1 \sigma_2 \sigma_3)
\end{aligned}
\label{eq:loss_learnable}
\end{equation}

where $\sigma_1, \sigma_2, \sigma_3$ are trainable parameters.

\textbf{Dataset and training:} We used \dyck-(5,*) datasets ($m=2, 5, 10, 20$). For each model on each dataset, the training is repeated five times with random initialization. 

\begin{figure}[h]
    \centering
    \includegraphics[width=\columnwidth]{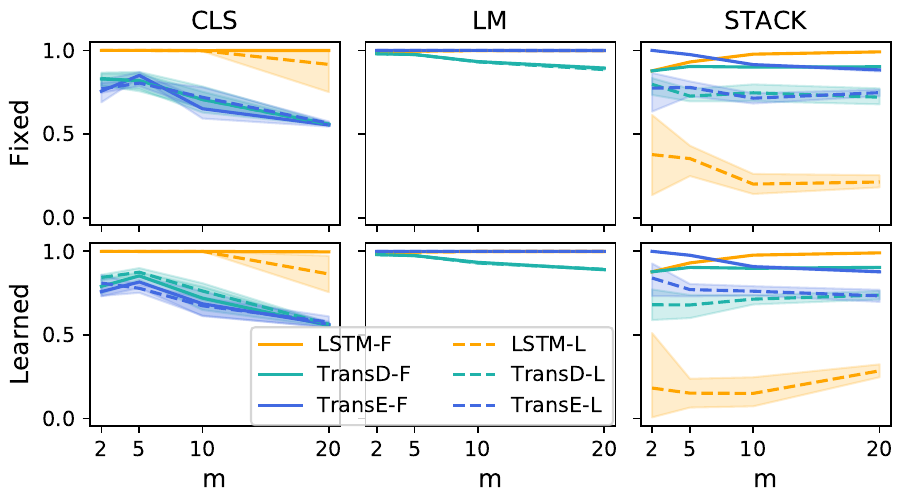}
    \caption{Model performance on \dyck-(5,*). First row shows results with Equation~(\ref{eq:loss}), and second row shows results with Equation~(\ref{eq:loss_learnable}). Lines indicate the average accuracy over 5 runs, and the shadows illustrate the range.}
    \label{fig:dyck_compare_loss}
\end{figure}

\textbf{Results:} Figure~\ref{fig:dyck_compare_loss} compares the model performance with fixed or learned weights and illustrates that neither the performance nor the hardship in factorization is sensitive to the choice of loss weights.

\section{Deeper models}
\begin{figure}[p]
\centering
    \begin{minipage}[b]{0.87\columnwidth}
    \centering
    \includegraphics[width=\columnwidth]{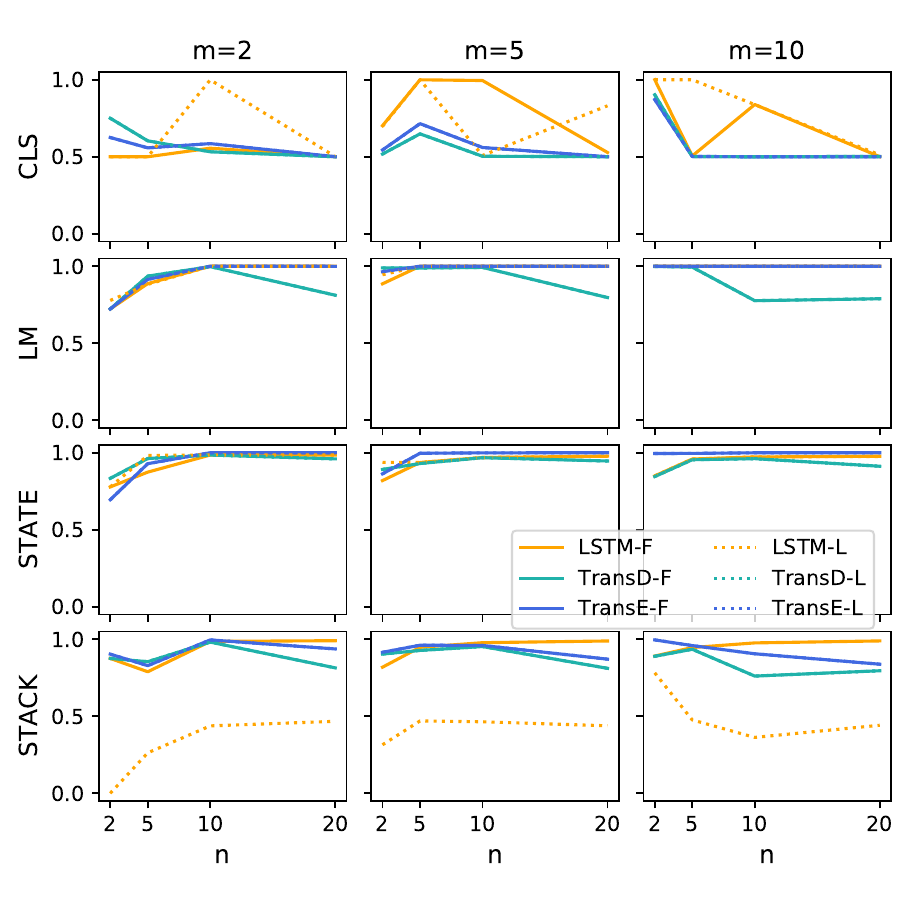}
    \end{minipage}
    \begin{minipage}[b]{0.87\columnwidth}
    \centering
    \includegraphics[width=\columnwidth]{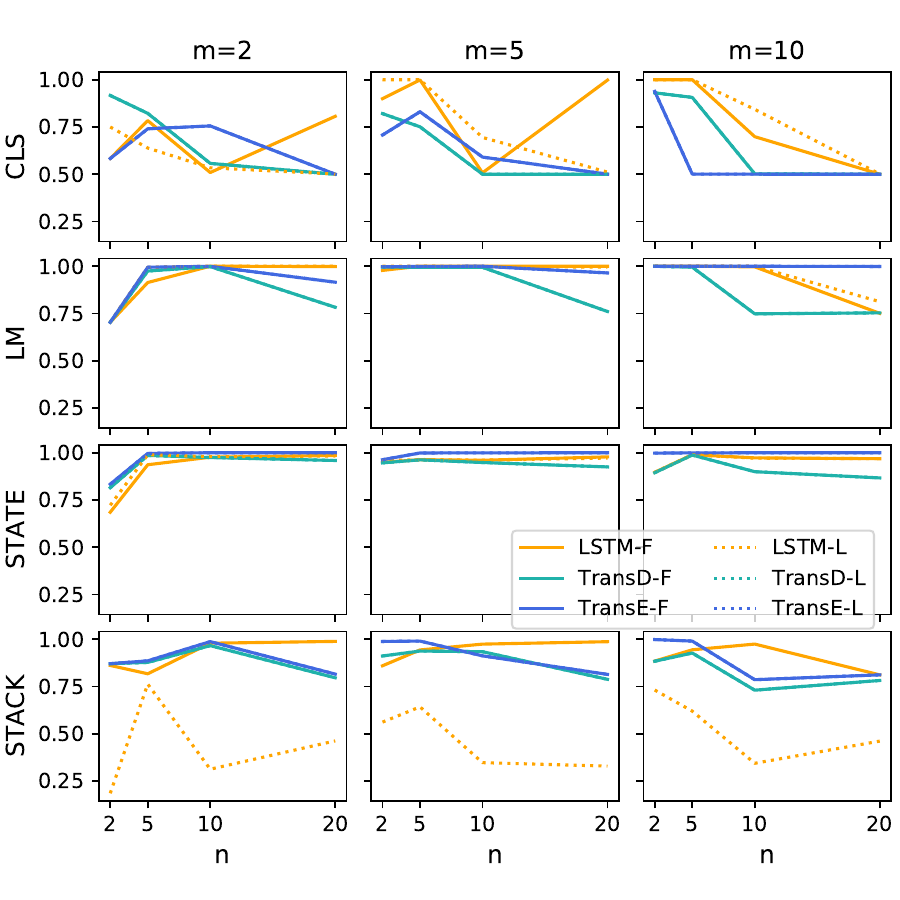}
    \end{minipage}
    \begin{minipage}[b]{0.87\columnwidth}
    \centering
    \includegraphics[width=\columnwidth]{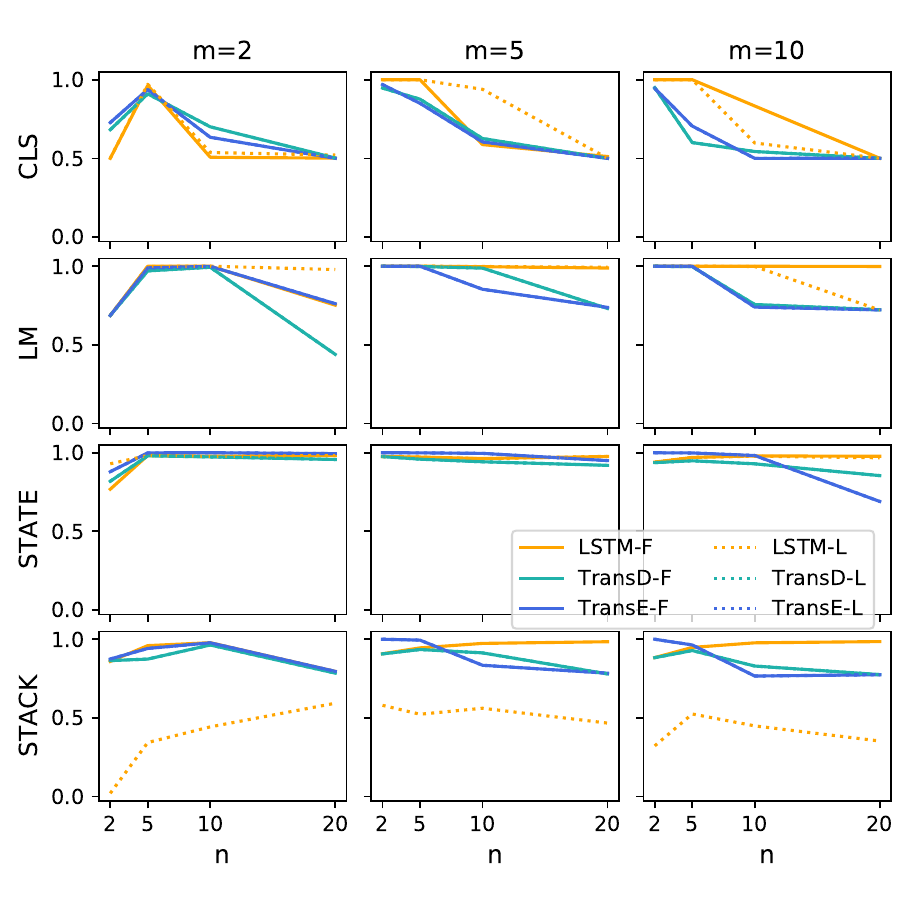}
    \end{minipage}
    \caption{4-layer model with $\alpha=4$ on (\wcwr)$^n$. The upper figure shows results on $k=3$, the middle shows $k=5$, and the bottom shows $k=10$.}
    \label{fig:wcwr_decom_s4}
\end{figure}

In Figure~\ref{fig:dyck_scale4},~\ref{fig:dyck_decom_4}, we show that additional layers and extra hidden size can not remedy the hardship in decomposition. Figure~\ref{fig:anbn_model_scale}, ~\ref{fig:wcwr_decom_s4} provide further evidence. From Figure~\ref{fig:anbn_model_scale}, we notice that the difference between forced and latent factorization is alleviated in Transformer models as the number of layer increases and $\alpha$ grows\footnote{The ablation on factorization is not conducted for L4S1 and L1S4 for any PDA. Also, since the factorization makes most significant difference on learning stack and Parity requires no stack, the comparison of factorization is not made on Parity.}. However, the gap remains huge for the LSTM model. Besides, the decrease of classification and language model accuracy is generally observed when scaling up the models, which indicates the training difficulty introduced by the deeper and wider model overwhelms the benefit of increased model capacity. 

\begin{figure}[t]
    \centering
    \includegraphics[width=\columnwidth]{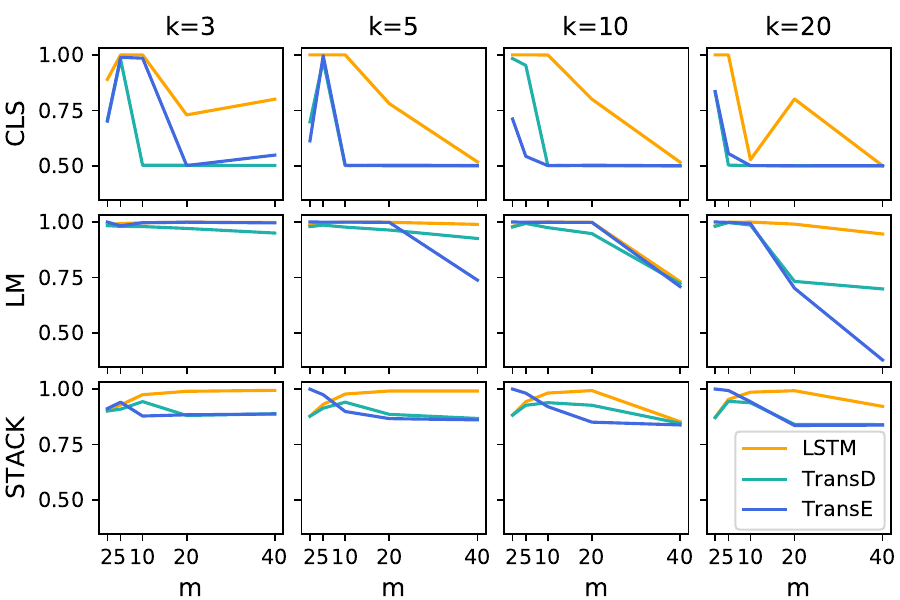}
     \caption{Performance on \dyck ($\alpha=1$, four layers)}
     \label{fig:dyck_s1l4}
\end{figure}     

\begin{figure}[t]
    \centering
    \includegraphics[width=\columnwidth]{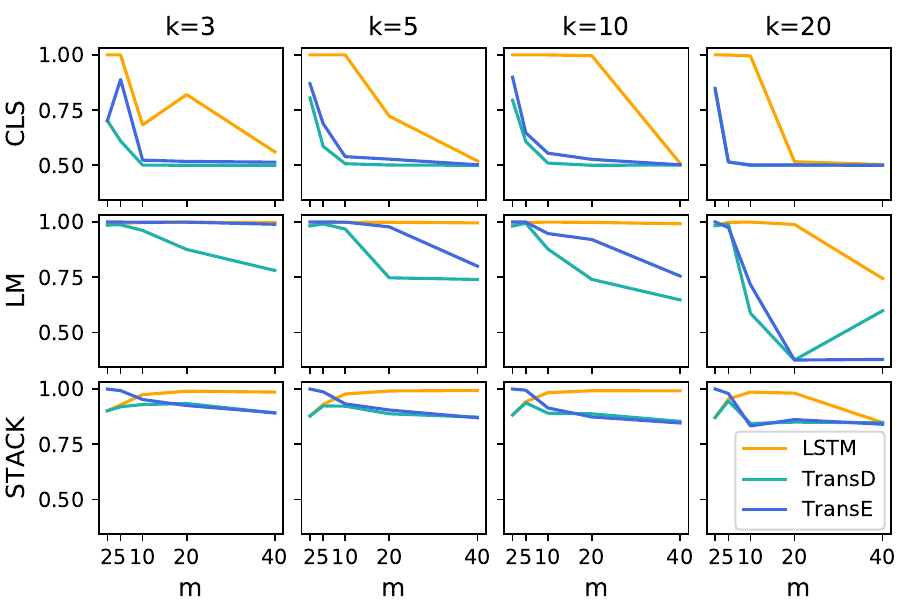}
    \caption{Performance on \dyck ($\alpha=4$, single layer)}
    \label{fig:dyck_s4l1}
\end{figure}

\end{document}